\documentclass[10pt,twocolumn,letterpaper]{article}

\usepackage[pagenumbers]{cvpr} 
\usepackage{soul}

\usepackage[dvipsnames]{xcolor}

\usepackage{multirow}
\usepackage{multicol}
\definecolor{cvprblue}{rgb}{0.21,0.49,0.74}
\usepackage[pagebackref,breaklinks,colorlinks,citecolor=cvprblue]{hyperref}

\def\confName{CVPR}
\def\confYear{2024}

\title{Closely Interactive Human Reconstruction with Proxemics and Physics-Guided Adaption}

\author{Buzhen Huang$^{1,2}$\footnotemark[1]\hspace{5mm} Chen Li$^{1}$\hspace{5mm} Chongyang Xu$^{3}$\hspace{5mm} Liang Pan$^{2}$\hspace{5mm} Yangang Wang$^{2}$\hspace{5mm} Gim Hee Lee$^{1}$\\%
\\
$^1$National University of Singapore \hspace{1mm}
$^2$Southeast University \hspace{1mm}
$^3$Sichuan University \hspace{1mm}\\
}

\begin{document}
\maketitle

\footnotetext[1]{The work was done while Buzhen Huang is a visiting student at National University of Singapore.}

\begin{abstract}
Existing multi-person human reconstruction approaches mainly focus on recovering accurate poses or avoiding penetration, but overlook the modeling of close interactions. In this work, we tackle the task of reconstructing closely interactive humans from a monocular video. The main challenge of this task comes from insufficient visual information caused by depth ambiguity and severe inter-person occlusion. In view of this, we propose to leverage knowledge from proxemic behavior and physics to compensate the lack of visual information. This is based on the observation that human interaction has specific patterns following the social proxemics. Specifically, we first design a latent representation based on Vector Quantised-Variational AutoEncoder (VQ-VAE) to model human interaction. A proxemics and physics guided diffusion model is then introduced to denoise the initial distribution. We design the diffusion model as dual branch with each branch representing one individual such that the interaction can be modeled via cross attention. With the learned priors of VQ-VAE and physical constraint as the additional information, our proposed approach is capable of estimating accurate poses that are also proxemics and physics plausible. Experimental results on Hi4D, 3DPW, and CHI3D demonstrate that our method outperforms existing approaches. The code is available at \url{https://github.com/boycehbz/HumanInteraction}.

\end{abstract}    

\section{Introduction}\label{sec:Introduction}
In everyday life, people continuously interact with each other to achieve goals or simply to exchange states of mind that enables us to live in groups and share skills and purposes. Analyzing human interactions in 3D space with computer vision techniques may promote the understanding of our social intellect. However, 
current multi-person human reconstruction methods often focus on pose accuracy~\cite{huang2022pose2uv,sun2021monocular,choi2022learning}, penetration avoidance~\cite{jiang2020coherent,zanfir2018monocular} or spatial distribution reasonableness~\cite{wen2023crowd3d,sun2022putting,huang2023reconstructing,huang2023crowdrec}, and always ignore the important close interactions. 

\begin{figure}
    \begin{center}
    \includegraphics[width=1.0\linewidth]{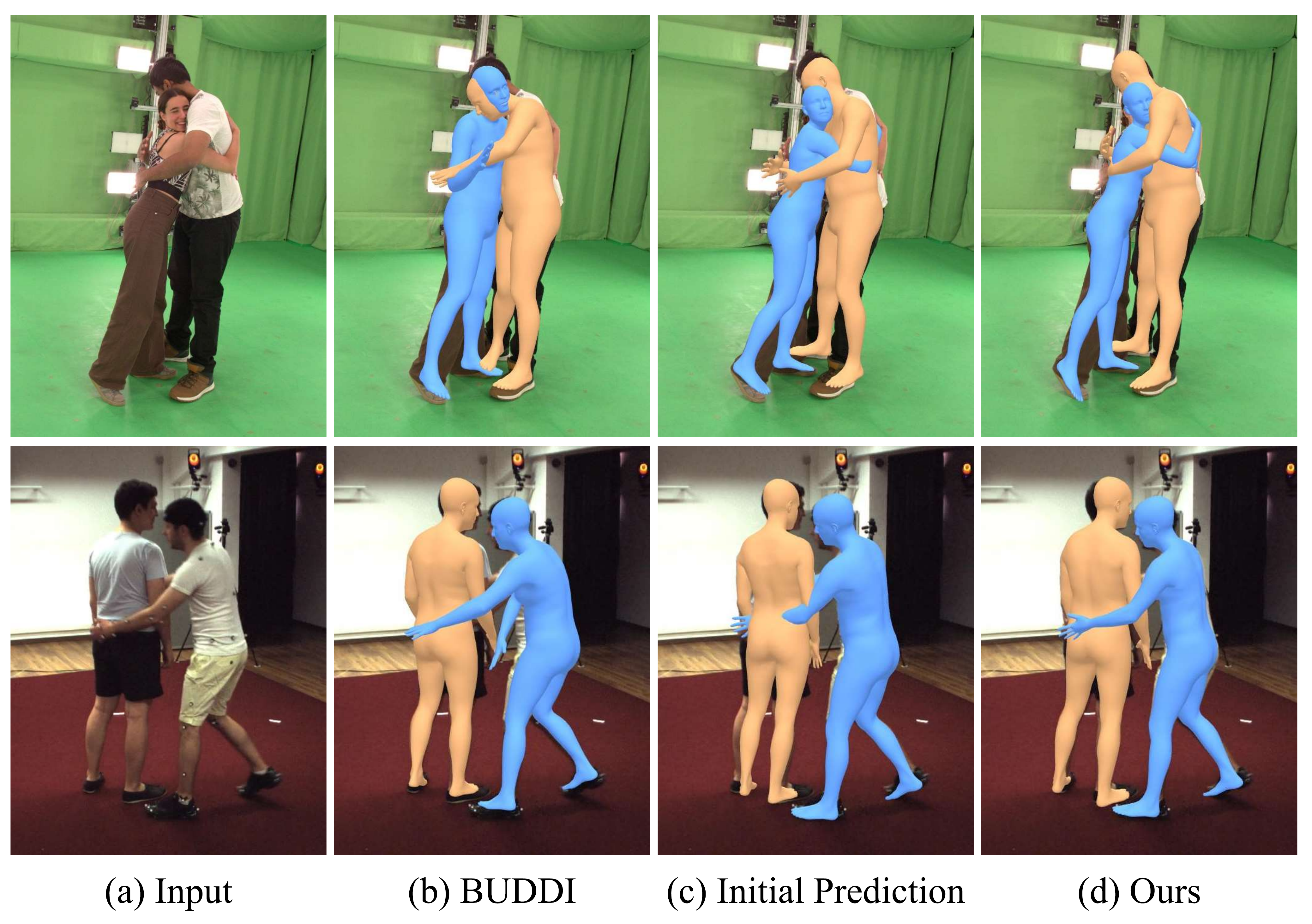}
    \end{center}
    \vspace{-7mm}
    \caption{Our method reconstructs closely interactive humans with plausible body poses, natural proxemic relationships and accurate physical contacts from single-view inputs. To address this challenging task, we formulate the reconstruction as a distribution adaption from the initial prediction~(c). Compared to the existing solution, BUDDI~\cite{muller2023generative}~(b), our method~(d) is more robust to visual ambiguity.}
\label{fig:teaser}
\vspace{-7mm}
\end{figure}

In this paper, we aim to reconstruct humans in close proximity with their physical social interactions from a monocular video. 
In addition to estimating plausible body poses, the task focuses on modeling natural interactive behaviors and accurate physical contacts.
The main challenge of this task comes from insufficient visual information caused by depth ambiguity and severe inter-person occlusion. This results in unsatisfactory prediction with high uncertainty. To circumvent the problems, we propose to leverage knowledge from proxemic behavior and physics to compensate the lack of visual information. This is based on the observation that human interaction has specific patterns following the social proxemics, and human can still infer plausible poses based on their prior knowledge even when there is severe occlusion. Inspired by this observation, we introduce a proxemics and physics guided diffusion model to reconstruct closely interactive humans.

First, we propose a latent representation to model human social interactions and proxemic behaviors. Historically, latent motion priors~\cite{shi2023phasemp,starke2022deepphase,ling2020character,rempe2021humor,zhang2021learning} have shown promising performance in modeling human dynamics. However, human interaction requires not only  vivid individual motions but also realistic interrelationships. We found that directly encoding the two-person interactions in a unified latent space with the similar structure as previous priors~\cite{ling2020character,zhang2021learning} compromise individual motion fidelity. We therefore design a dual-branch network to learn two discrete codebooks to model the social interactions with Vector Quantised-Variational AutoEncoder~(VQ-VAE)~\cite{van2017neural}. These codebooks represent high-fidelity motions and can share information during the inference phase. 

Subsequently, we design a diffusion model to reconstruct 3D human interactions from monocular videos. However, conventional diffusion models always start from the standard Gaussian distribution to predict a clean sample, and the early denoising iterations primarily produce noises with limited information~\cite{yuan2023physdiff,xu2023interdiff}. Therefore, we first regress an initial distribution from image features to draw plausible poses for the denoising process. Although the coarse poses may not be fully consistent with image observations due to the visual ambiguity and occlusion, they are valid signals to be evaluated by the latent interaction prior and physical constraints. 
We thus use the diffusion model as an adaptor to refine the initial distribution under the guidance of learned interaction prior, physical constraints, and image observations. In each timestep, the current interactive poses are fed into the interaction prior to find the closest pair, which are then used to update the current states. 
In addition, we further design a metric to evaluate the penetration, which reflects the physical plausibility of the current interaction state. We also project the 3D joint positions to the 2D image plane and calculate the projection loss gradients, and enforce the results to be consistent with the image observation. The physical loss, projection loss gradients, and image features are then concatenated as a condition to guide the diffusion model to achieve the distribution adaption. After several diffusion timesteps, we can obtain the final distribution to draw accurate and realistic 3D interactions. To summarize, the main contributions of this paper are as follows:
\begin{itemize}
  \item We formulate the close interaction reconstruction as a distribution adaption process, which can estimate plausible body poses, natural proxemic relationships and accurate physical contacts from a single-view video.

  \item We design a dual-branch discrete prior to learn human interactive behaviours that can provide additional knowledge for single-view reconstruction.

  \item We propose a novel diffusion model to incorporate proxemics, physics, and image observations for improving the reconstruction. This model achieves state-of-the-art performance in closely interactive scenarios.
  
\end{itemize}

\begin{figure*}
    \begin{center}
    \includegraphics[width=1.0\linewidth]{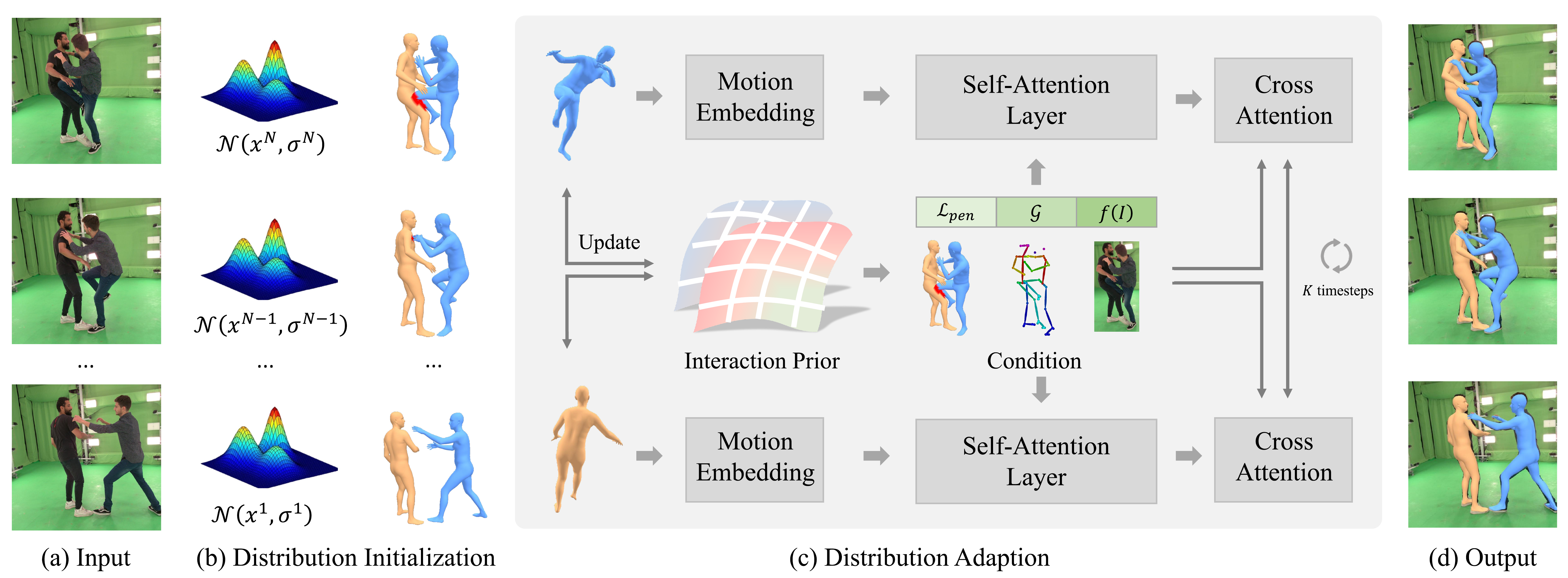}
    \end{center}
    \vspace{-6mm}
    \caption{\textbf{Overview of our method.} Given a monocular video with interactive humans~(a), we first regress an initial distribution for each person (Inter-person penetration is marked in \textcolor{red}{red})~(b). To refine the distribution, we design a proxemics and physics guided diffusion model to achieve distribution adaption~(c). Specifically, the motions drawn from the initial distributions are updated by a discrete interaction prior. The updated motions are then fed into a dual-branch diffusion model to denoise under the guidance of physics and image observations. The denoised motions are then used as the input for the next timestep. The adaption takes several diffusion timesteps and finally produce accurate results~(d).}
\label{fig:pipeline}
\vspace{-6mm}
\end{figure*}

\section{Related Work}\label{sec:relatedwork}
\noindent\textbf{Single-person shape and pose estimation.}
Early works in single-person shape and pose estimation directly predict a 3D mesh with optimization~\cite{Bogo:ECCV:2016} or regression~\cite{kanazawa2018end}. With the development of computer vision techniques, more advanced backbones~\cite{sun2019deep,dosovitskiy2020image,goel2023humans}, representations~\cite{li2021hybrik,huang2022object}, camera models~\cite{li2022cliff}, and priors~\cite{SMPL-X:2019} are introduced to improve the estimation. To exploit the temporal context of human motion for better temporal consistency, video-based methods use the recurrent neural network~\cite{kocabas2019vibe} or transformer~\cite{zheng20213d} to process the input. However, these methods encounter global inconsistency, since they can only produce root-relative motion. Since the aforementioned methods do not consider depth ambiguity and occlusion from monocular motion capture, recent diffusion-based works~\cite{gong2023diffpose,choi2022diffupose,holmquist2023diffpose} model the uncertainty of 2D-3D lifting for 3D pose estimation. Nevertheless, while single-person methods can achieve better performance in joint accuracy, they do not consider the relationships between humans in the same scene and cannot be used to reconstruct human interactions.

\vspace{-3mm}
\paragraph{Multi-person shape and pose estimation.} 
Recently, along with the success of deep learning in 3D computer vision, multi-person shape and pose estimation has made tremendous progress. To address the inter-person occlusions, some works~\cite{qiu2023psvt,li2023coordinate,sun2021monocular,choi2022learning} design various model architectures to extract valid features and can recover accurate body meshes from monocular images. However, they do not consider absolute positions and cannot model delicate human interactions. To estimate multi-person meshes in a unified 3D space, a few works estimate absolute translations with projection geometry~\cite{cha2022multi,zanfir2018deep,huang2022pose2uv,zanfir2018monocular}, but the strategy is affected by depth and body shape coupling~\cite{ugrinovic2021body}. In addition, it also strongly depends on the quality of 2D poses, which is hard to obtain in interactive cases. Recently, a few works introduce position representation~\cite{sun2022putting,zhang2021body}, 6D pose estimation~\cite{mustafa2021multi}, and ordering-aware loss~\cite{jiang2020coherent,khirodkar2022occluded,huang2023reconstructing,wen2023crowd3d} to constrain relative positions among humans. Nonetheless, the coarse ordinal relation is inadequate for close interaction reconstruction. Only a few works explicitly consider the close interactions. Fieraru~\etal designed several losses to prevent self-collision and interpenetration. \cite{fieraru2020three,fieraru2023reconstructing} incorporate additional contact constraints in the optimization. BUDDI~\cite{muller2023generative} further trains a diffusion-based proxemics prior to assist the fitting. However, all these methods cannot utilize temporal information and are still confronted with visual ambiguities.

\vspace{-3mm}
\paragraph{Closely interactive motion generation.}
Human motion generation is a hot topic in recent years. Different from single person scenarios~\cite{tevet2022human}, the closely interactive behavior is difficult to record, and the insufficient data is the bottleneck of multi-person motion generation for a long time. Early work~\cite{yin2018sampling} can only adopt sampling strategy to generate interactions. Due to the scarcity of training data, some recent works~\cite{won2021control,zhang2023simulation} use single-person data with reinforcement learning to achieve multi-person motion generation. The trained models cannot be generalized to various interaction types~\cite{shafir2023human,won2021control}. Other works~\cite{peng2023trajectory,tanke2023social,xu2022stochastic,xu2023joint} train the models with motion capture data, but the data contains only a few close interaction cases. Game engines can also be used to generate training data for close interactions~\cite{starke2021neural,xu2023actformer,starke2020local}. With the development of motion capture techniques, the recent works can use real interaction data to generate human reactions~\cite{guo2022multi,fang2023pgformer,rahman2023best,chopin2023interaction}. InterGen~\cite{liang2023intergen} builds a large-scale two-person interaction dataset based on a system with massive cameras, which promote the close interaction generation. It also proposes a diffusion-based network to generate realistic interactions. However, all the above methods adopt skeletons to represent the interactive pairs, and only focus on the pose accuracy. 
Furthermore, although they can generate interactive behaviors, their models cannot be used as a prior for other downstream tasks~(\eg, monocular motion capture). In contrast, our interaction prior based on mesh representation can consider the important body contacts.

\section{Our Method}\label{sec:method}
In this work, we aim to reconstruct closely interactive human motions from monocular videos. Since it is difficult to obtain sufficient valid information from single-view images, we design a discrete interaction representation to learn interactive human behaviors to assist the motion capture. We then model the uncertainty of human interaction as a set of Gaussian distributions and formulate the interaction reconstruction as a distribution adaption. With a dual-branch diffusion model, we can gradually adapt interaction distributions under the guidance of proxemic behaviors, physical laws and image observations.

\subsection{Representation}\label{sec:Representation}
For an interactive pair, we use two SMPL models~\cite{loper2015smpl} to represent the interactions. We also adopt 6D representation~\cite{zhou2019continuity} to describe joint rotations, and thus the parameters for a single person $x = \left\{\theta, \beta, \tau\right\}$ consist of pose $\theta \in \mathbb{R}^{144}$, shape $\beta \in \mathbb{R}^{10}$ and translation $\tau \in \mathbb{R}^{3}$. An interactive motion with $N$ frames of two people can be denoted as $\mathbf{x}^{1:N} = \left\{\mathbf{x}^{a,1:N}, \mathbf{x}^{b,1:N} \right\}$, where $\mathbf{x}^{a,1:N} = \left\{x^i\right\}_{i=1}^N$. We use $\hat{\mathbf{x}}$ to represent ground-truth data. Since the single-view reconstruction is an ill-posed problem, we use a set of Gaussian distributions $\mathcal{N} \left(\mathbf{x}^{1:N}, \sigma^{1:N}\right)$ to model the uncertainty, where $\sigma$ is the variances. A motion represented with SMPL parameters can be directly drawn from the distributions. 
Given a video $\mathcal{I} = \left\{I^i\right\}_{i=1}^N$ with close interactions of two people, the output of our network are two sets of distributions $\left\{\mathcal{N}^a, \mathcal{N}^b\right\}$.

\subsection{Discrete Interaction Prior}\label{sec:prior}
The way that people react to and interact with the surrounding world is a result of evolution that follows certain rules.
With only partial observations from single-view images, humans can also infer the complete 3D poses for an interactive pair. 
We thus hope to learn the interactive behaviors to provide additional knowledge for the monocular interaction reconstruction. Recently, VQ-VAE~\cite{van2017neural} achieves promising performance in modeling prior knowledge across different modalities~\cite{razavi2019generating,zhang2023t2m}. In addition, the discrete representation can be used to measure the certain distance between the predicted results from the prior, which is a significant strength over common VAEs~\cite{kingma2013auto,zuo2023reconstructing} in reconstruction tasks. We therefore design a novel dual-branch VQ-VAE model for modeling the two-person interactions. 

\begin{figure}
    \begin{center}
    \includegraphics[width=1.0\linewidth]{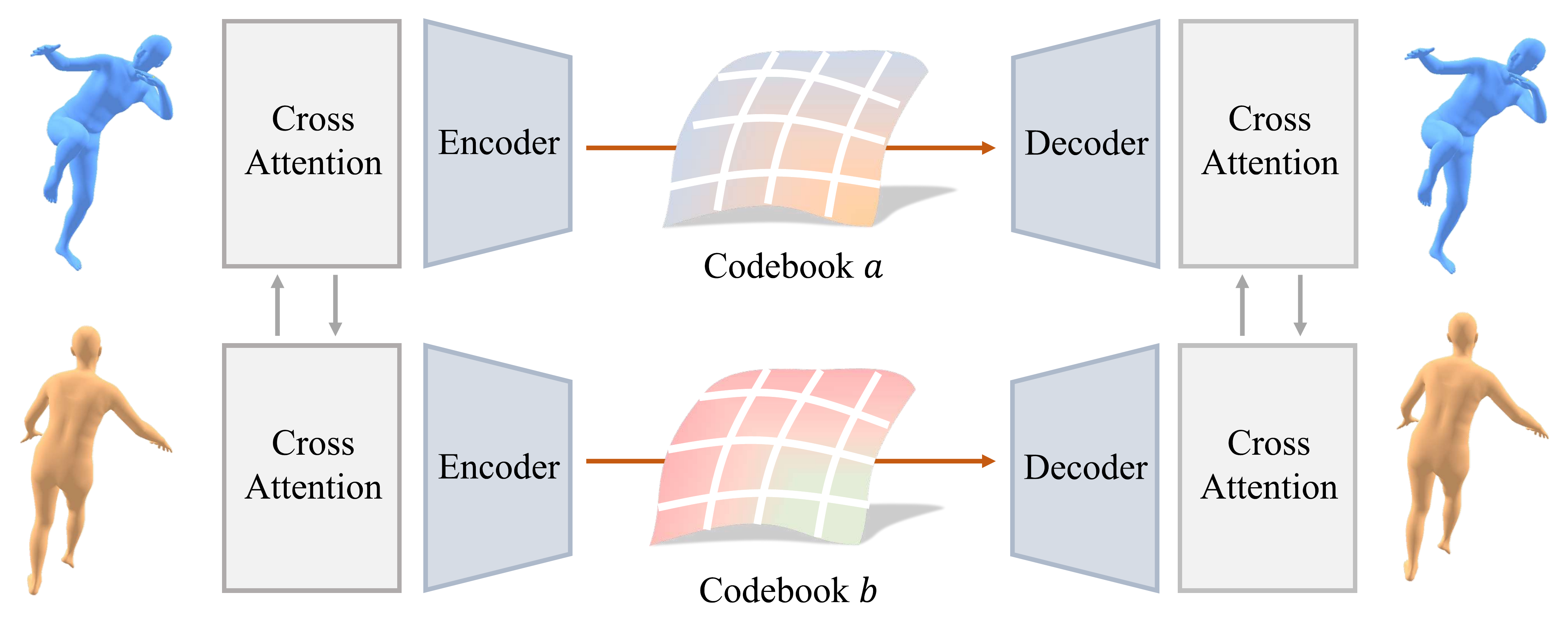}
    \end{center}
    \vspace{-6mm}
    \caption{The prior has a dual-branch structure, with each branch representing the motion of a character. Each branch has a codebook learned by VQ-VAE, which models interactive behaviours. In addition to the codebook, the two branches share the same weights, and they can exchange information with the cross-attention module.}
\label{fig:prior}
\vspace{-6mm}
\end{figure}

\vspace{-3mm}
\paragraph{Motion state.} 
We found that joint position and velocity are important to learn the interactive behaviour. In addition to the SMPL parameters $\{\theta, \beta, \tau\}$, we further combine the joint positions $J$, joint velocities $\dot{J}$, and joint positions relative to the counterpart $J^\prime$ in to the motion states for learning the prior. The motion state of a character is denoted as $\mathbf{X} = \left\{\theta, \beta, \tau, J, \dot{J}, J^\prime \right\}$.

\vspace{-3mm}
\paragraph{Model architecture.} 
Since directly encoding the two-person interactions in a unified latent space with the similar structure as previous priors~\cite{ling2020character,zhang2021learning} compromise individual motion fidelity, we thus design a dual-branch network to learn the discrete prior. As shown in \cref{fig:prior}, the two branches represent two individuals respectively and can share information with each other. Specifically, the motion states of two characters are first passed through a motion embedding layer to obtain the latent state $h_0$. We then add a positional encoding on the encoded latent state, and feed it into $S$ transformer blocks to obtain denoised hidden states $h_{S}$. Each block consists of two multi-head attention layers followed by one feed-forward network. In the first attention layer, the individual state $h_s^a$ and condition $c$ are first processed with an adaptive layer norm~\cite{peebles2023scalable}, and then the normalized state $\bar{h}_s^a$ go through a multi-head self-attention module:
\begin{equation}
    \mathbf{c}^a_{s}=\operatorname{Attn}(\mathbf{Q}, \mathbf{K}, \mathbf{V}),
\end{equation}
where we set the $\mathbf{Q}, \mathbf{K}, \mathbf{V}$ to be:
\begin{equation}
    \mathbf{Q}=\bar{h}_s^a \mathbf{W}_{s1}^Q, \quad \mathbf{K}=\bar{h}_s^a \mathbf{W}_{s1}^K, \quad \mathbf{V}=\bar{h}_s^a \mathbf{W}_{s1}^V.
\end{equation}
$\mathbf{W}^Q$, $\mathbf{W}^K$ and $\mathbf{W}^V$ are projection weights for query, key and value.

The second attention layer has the same structure as the first, 
but the key and value are the states of the counterpart: 
\begin{equation}
    \mathbf{Q}=\bar{h}_s^a \mathbf{W}_{s2}^Q, \quad \mathbf{K}=\bar{h}_s^b \mathbf{W}_{s2}^K, \quad \mathbf{V}=\bar{h}_s^b \mathbf{W}_{s2}^V.
\end{equation}
With this design, one branch can sense the states of the other and generate more plausible interactions. A feed-forward layer is then used to regress the states in the next stage from the output of two attention layers.

With the above encoder, the motion states are mapped to a set of latent codes $\mathbf{Z} = E(\mathbf{X})$. For each latent code $z_i \subset \mathbf{Z}$, we find the closest vectors from corresponding codebook $C$ by calculating the Euclidean distance, \ie: 
\begin{equation}
\hat{z}_i=\underset{c_k \in C}{\arg \min }\left\|z_i-c_k\right\|_2.
\end{equation}
A decoder with the same structure as the encoder is then used to reconstruct the states from sampled latent codes $\hat{\mathbf{Z}}$. In addition to the codebooks, the two branches share the same weights. 

\vspace{-3mm}
\paragraph{Model training.} 
To train the VQ-VAE, we optimize the following objectives:
\begin{equation}
\mathcal{L}_{v q}=\mathcal{L}_{rec}+\alpha \underbrace{\|Z-\operatorname{sg}[\hat{Z}]\|_2}_{\mathcal{L}_{\text {commit }}},
\label{eq:Lvq}
\end{equation}
where $\operatorname{sg}[\cdot]$ is the stop-gradient operator and $\alpha$ is a hyper-parameter for the commitment loss $\mathcal{L}_{\text{commit}}$. The reconstruction loss is given by:
\begin{equation}
\mathcal{L}_{rec}=\mathcal{L}_{smpl}+\mathcal{L}_{joint}+\mathcal{L}_{vel}+\mathcal{L}_{int},
\label{eq:Lrec}
\end{equation}
which is the sum of the supervisions from the SMPL parameters: $\mathcal{L}_{\text{smpl}} = \| [\beta, \theta] - [\hat{\beta}, \hat{\theta}] \|_2^2$, 3D joint positions: $\mathcal{L}_{\text{joint}} = \| J_{3D} - \hat{J_{3D}} \|_2^2$ and velocities: $\mathcal{L}_{\text{vel}} = \| \dot{J}_{3D} - \hat{\dot{J}}_{3D} \|_2^2$ 
on each character.

We also supervise the relative distance between two characters, \ie:
\begin{equation}
    \mathcal{L}_{\text{int}} = \| |J^a_{3D} - J^b_{3D}| - |\hat{J}^a_{3D} - \hat{J}^b_{3D}| \|_2^2.
\end{equation}
To avoid codebook collapse~\cite{razavi2019generating}, the cookbooks are optimized by exponential moving average~(EMA) and codebook reset~(Code Reset) operations following \cite{zhang2023t2m}.

\subsection{Initial Interaction Distribution Prediction}\label{sec:prediction}
To achieve the reconstruction, we use diffusion model as a distribution adaptor, which has shown its powerful performance in generative tasks~\cite{rombach2022high,yuan2023physdiff}. 
However, conventional diffusion model always start from standard Gaussian noises and the early denoising iterations primarily produce noises with limited information~\cite{yuan2023physdiff,xu2023interdiff}.
Previous works thus require more diffusion timesteps to generate satisfactory results. Different from text and other high-level signals, images contain more prior knowledge and can relieve the uncertainty. 

We thus first regress initial distributions from images for an interactive motion. To break the constraints of limited interaction data and improve the generalization ability, we design the initial distribution prediction network to have a single-person structure. Specifically, we adopt a ViT~\cite{dosovitskiy2020image} to extract image features for a single person, and then combine the image features and bounding-box information to regress SMPL parameters with a transformer decoder. To obtain absolute positions for representing interaction, the translation $\tau$ of SMPL model is transformed from estimated camera parameters like CLIFF~\cite{li2022cliff}.
Subsequently, we train the network on single-person pose estimation datasets with the following loss functions:
\begin{equation}
\mathcal{L}_{init}=\mathcal{L}_{smpl}+\mathcal{L}_{joint}+\mathcal{L}_{reproj},
\end{equation}
where $\mathcal{L}_{smpl}$ and $\mathcal{L}_{joint}$ are the SMPL and 3D joint position supervisions defined after Eq.~\ref{eq:Lrec}. The reprojection loss is given by:
\begin{equation}
    \mathcal{L}_{\text{reproj}} = \| \Pi\left(J_{3 D}+ \tau\right) - \hat{J_{2D}} \|_2^2,\label{equ:reproj}
\end{equation}
where $\Pi(\cdot)$ projects the 3D joints to 2D image with camera parameters, and $\hat{J_{2D}}$ is ground-truth 2D pose.

Although we can use sufficient data to train the single-person network, it still cannot work well on closely interactive cases. The reason is that the current deep learning algorithms cannot identify the belonging of each pixel in complex interaction cases, which induces severe visual ambiguity for the reconstruction. In addition, the inter-person occlusions further increase the uncertainty. 
We thus use the predicted SMPL parameters $x$ to construct the initial distributions $q \left(x, \sigma\right)$ for the diffusion model. The coarse distribution is then used in the diffusion model to achieve the adaption.

\subsection{Interaction Distribution Adaption}\label{sec:adaption}
With the initial values, we first associate each person across time with predicted poses, positions and 2D bounding-boxes. The procedure is similar to PHALP~\cite{rajasegaran2022tracking}. The difference is that we do not consider the appearance due to the serious overlapping in the interactive image. After the association, we can obtain two sets of Gaussian distributions $\left\{\mathcal{N}^{a, 1:N}, \mathcal{N}^{b,1:N}\right\}$. 

\vspace{-3mm}
\paragraph{Dual-branch diffusion model.}
We then design a dual-branch diffusion network to refine the initial distributions. As shown in \cref{fig:pipeline}, the diffusion model has the same structure as the interaction prior encoder. After the $S$ transformer blocks, we concatenate the hidden states $h_{H}$ and bounding-box information to regress pose, shape and translation parameters as in \cref{sec:prediction}. The output parameters are then transformed into distributions for the next diffusion timestep.

\vspace{-3mm}
\paragraph{Diffusion with initial distributions.}
To train the diffusion model for pose estimation, previous methods~\cite{rommel2023diffhpe,holmquist2023diffpose} inject time-dependent noises sampled from standard Gaussian distribution to ground-truth motion $\hat{\mathbf{x}}_0$, which can be formulated as:
\begin{equation}\label{equation:forward_diffusion_origin}
    q(\mathbf{x}_t \mid \hat{\mathbf{x}}_0) = \sqrt{\hat{\alpha}_t}\hat{\mathbf{x}}_0 + \sqrt{1 - \hat{\alpha}_t} \epsilon, \epsilon \sim \mathcal{N} \left(0, \rm{I} \right),
\end{equation}
where $\alpha_t$ is a constant hyper-parameter~\cite{nichol2021improved}, and $\hat{\alpha}_t = \prod_{i=0}^t \alpha_i$.

We found that $\mathbf{x}_t$ follows standard Gaussian distribution and the early iterations are meaningless for a human motion, which results in the model to spend more timesteps in the reverse diffusion process. 
We thus diffuse towards the initial distributions. It should be noted that directly sample noises from the initial distributions will result in a deviant distribution $x_t \sim \mathcal{N} \left(Tx, \sigma \right)$ since $x$ is not equal to zero. Consequently, we change the diffusion process to be:
\begin{equation}\label{equation:forward_diffusion}
    q(\mathbf{x}_t \mid \hat{\mathbf{x}}_0) = \mathbf{x} + \sqrt{\hat{\alpha}_t} (\hat{\mathbf{x}}_0 - \mathbf{x}) + \sqrt{1 - \hat{\alpha}_t} \epsilon, \epsilon \sim \mathcal{N} \left(0, \sigma \right).
\end{equation}

The sampled motions that satisfied the discrete interaction prior and physical laws are considered plausible, and the output signals can be used to guide the denoising process.
With the above diffusion process, a generative model can be obtained by reversing the process, starting from samples $\mathbf{x}_t \sim \mathcal{N} \left(\mathbf{x}, \sigma \right)$, which is defined as:
\begin{equation}\label{equation:reverse_diffusion}
    q(\mathbf{x}_{t-1} \mid \mathbf{x}_{t},c) = \mathcal{N} (\mathbf{x}_{t-1}; \mu_{\alpha}(\mathbf{x}_{t},c),\tilde{\beta_t} \sigma),
\end{equation}
where $\mu_{\alpha}(\mathbf{x}_{t},c)$ is the estimated mean by the diffusion model under the condition of $c$ in timestep $t-1$. $\tilde{\beta_t}$ is the variance calculated by the hyper-parameters $\beta_t$, $\hat{\alpha}_t$ and $\hat{\alpha}_{t-1}$.

\vspace{-3mm}
\paragraph{Proxemics and physics guided denoising.}
With the diffusion framework, we can gradually refine the initial distributions in each timestep under the condition of image features~\cite{gong2023diffpose}. However, the image features in interactive cases are always ambiguous, and the model cannot get sufficient information to generate a valid interaction. 
We design discrete interaction prior to provide additional knowledge for the distribution adaption. 

For a sampled interactive motion, we first calculate the character states in each frame and then feed the states to the trained interaction prior. The states of two characters are then encoded to two latent codes by the interaction prior encoder. We can find the closest codes from the codebooks by calculating the Euclidean distance between the encoded latent codes and codebooks. We then recombine the closest latent codes in the codebooks and decode them to a new interaction. By this way, although the initial motion may not be a valid interaction due to the visual ambiguity and occlusion, we can enforce the motion to follow proxemic and interactive behaviors with the trained interaction prior. The output motions from the prior are then used as the input of diffusion model. 

However, the prior only considers the interactive behavior, and cannot guarantee the physical plausibility. Since body contacts are important for human interactions, we further use a physical metric to evaluate the penetration on two-person meshes. Specifically, for the interactive meshes output from the prior, we first detect the set of colliding triangles using bounding volume hierarchies~(BVH)~\cite{Karras2383801}. 
The penetrated vertices are then used to sample the values in local 3D distance fields~\cite{Tzionas2016}, which reflect the degree of mesh penetrations and collisions. The penetration loss is therefore written as:
\begin{equation}
    \begin{array}{r}
    \mathcal{L}_{pen}=\sum \limits_{\left(f_a, f_b\right) \in \mathcal{C}}\left\{\sum_{v_a \in f_a}\left\|-\Psi_{f_b}\left(v_a\right) n_a\right\|^2+\right. \\
    \\
    \left.\sum_{v_b \in f_b}\left\|-\Psi_{f_a}\left(v_b\right) n_b\right\|^2\right\},
\end{array}
\end{equation}
where $f_a, f_b$ are two colliding triangles in the detected colliding triangles $\mathcal{C}$. $v$ and $n$ are vertex position and normal, respectively, and $\Psi(\cdot)$ is the distance field. The values are summed as a condition in the reverse diffusion process.

We also require the interaction to be consistent with the image observations. For 3D poses in the current timestep, we calculate the gradient vector of 3D joints to detected 2D keypoints: 
\begin{equation}
    \mathcal{G} = \frac{\partial \left\| \Pi\left(J_{3 D}\right) - p_{2D} \right\|_2^2}{\partial J_{3 D}}.
\end{equation}

Finally, the condition $c$ of the diffusion model combines the image features, penetration values $\mathcal{L}_{pen}$, and projection loss gradients $\mathcal{G}$.

\vspace{-3mm}
\paragraph{Model training.}
We train the diffusion model with the following loss functions:
\begin{equation}
    \mathcal{L} =  \mathcal{L}_{\text{reproj}} + \mathcal{L}_{\text{smpl}} + \mathcal{L}_{\text{joint}} + \mathcal{L}_{\text{vel}} + \mathcal{L}_{\text{int}} + \mathcal{L}_{\text{pen}},\label{equ:loss}
\end{equation}
where the loss terms are defined in previous sections. 
In addition, we randomly mask image features and projection loss gradients in the condition to address the missing detections and visual ambiguities.

\section{Experiments}\label{sec:Experiments}

\begin{figure*}
    \begin{center}
    \includegraphics[width=1.0\linewidth]{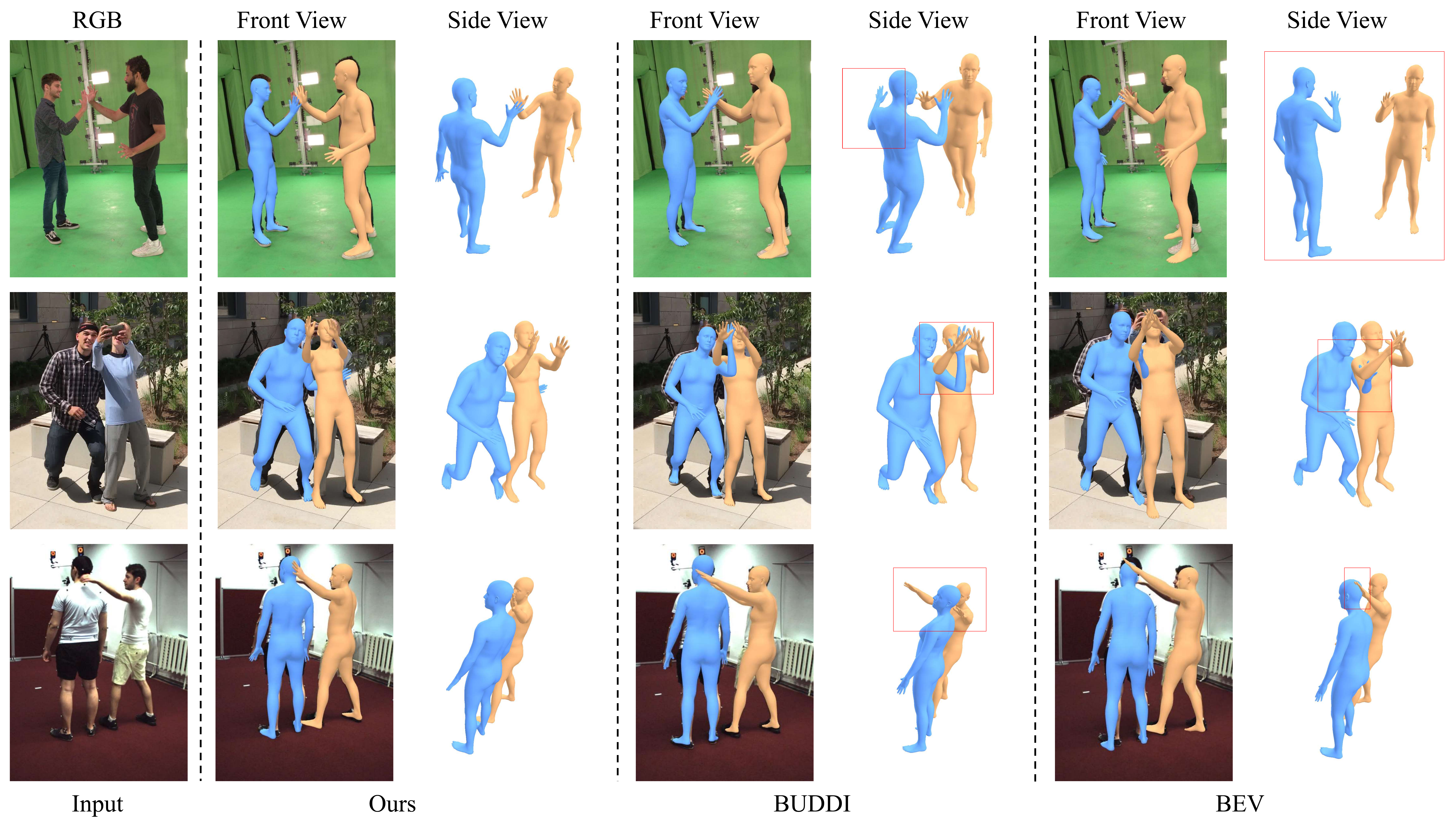}
    \end{center}
    \vspace{-6mm}
    \caption{Qualitative comparison with BUDDI~\cite{muller2023generative} and BEV~\cite{sun2022putting}. Our method can produce more accurate body poses and physical contacts.}
\label{fig:comparison}
\vspace{-4mm}
\end{figure*}

\subsection{Datasets}\label{sec:datasets}
\noindent\textbf{Common datasets}. Following previous work, the typical datasets 
the typical datasets used for training are Human3.6M~\cite{h36m_pami}, MPI-INF-3DHP~\cite{mehta2017monocular}, COCO~\cite{lin2014microsoft}, MPII~\cite{andriluka20142d}. We use the image and video datasets to train the initial dsitribution prediction network. The video datasets are also used to train the diffusion model by masking the features from the counterpart. 
\noindent\textbf{InterHuman~\cite{liang2023intergen}} is a large-scale dataset containing diverse two-person interactions. Due to the lack of color images, we use it to train the discrete interaction prior only.
\noindent\textbf{Hi4D~\cite{yin2023hi4d}} is an accurate multi-view dataset for closely interacting humans. It contains 20 unique pairs of participants with varying body shapes and clothing styles to perform diverse interaction motion sequences. 5 pairs~(23, 27, 28, 32, 37) are used as testset, and the rest are used for training.
\noindent\textbf{CHI3D~\cite{fieraru2020three}} captures 3 pairs of people in close interaction scenarios with a Vicon MoCap system and 4 additional RGB cameras. We use the standard splits of this dataset.
\noindent\textbf{3DPW~\cite{vonMarcard2018}} also contains several sequences with two-person interactions. We use these sequences for evaluation only.

\subsection{Metrics}\label{sec:Metrics}
We report the Mean Per Joint Position Error (MPJPE), and 
MPJPE after rigid alignment of the prediction with ground truth using Procrustes Analysis (PA-MPJPE) on these datasets. To measure the mesh quality, the Mean Per Vertex Position Error (MPVPE) is also used. In addition, we further evaluate the interaction error, which is defined as $\footnotesize{\frac{1}{K\times K} \sum\limits_{i=1}^K\sum\limits_{j=1}^K|\|J^a_i - J^b_j\| - \|\hat{J}^a_i - \hat{J}^b_j\||}$, and $K$ is the number of joints.

\subsection{Comparison to State-of-the-Art Methods}\label{sec:Comparison}
We conduct several experiments to demonstrate the effectiveness of our method in closely interactive scenarios. We choose the current SOTA single-person and multi-person mesh recovery models as baseline methods. Human4D and CLLIF are designed for single-person scenarios. They can achieve high joint accuracy, but do not consider the interactions. BEV and GroupRec aim to multi-person cases and incorporate spatial constraints into the reconstruction. In addition, BUDDI is the most relevant work to ours, which reconstruct interactive humans from single images with a optimization. \cref{tab:Hi4D} shows a quantitative comparison with these baseline methods on Hi4D dataset. All methods are finetuned on Hi4D training set for a fair comparison. Since Human4D uses a weak-perspective camera model, it cannot obtain accurate absolute positions in camera coordinates. Although Human4D can predict precise root-relative joint positions, it cannot be used to reconstruct interactive pairs. Different from Human4D, CLIFF estimates humans in the original camera coordinates and can produce valid relative positions for interactive humans. However, it processes each human iteratively and results in a high interaction error due to depth ambiguity. We also compare our method with BEV and GroupRec. They constrain the spatial positions of humans, but the constraint can only work well on relative large scenes. Besides, BUDDI trains a proxemic prior to fit human models to interactive images. However, the optimziation is not robust to depth ambiguity and 2D pose noises. In \cref{fig:comparison}, we find that BUDDI overfits to the noisy 2D poses and produce a wrong result. Although it is designed for close interaction reconstruction, its performance is strongly affected by the 2D pose detection. Thus, it is still inferior to other baseline methods. In addition, its optimization framework is time-consuming, which takes 3 $\sim$ 4 min to reconstruct a pair. In contrast, our method takes the detected 2D poses as one of conditions and is more robust to detection noises due to the random masking strategy. With the assistance of the discrete prior, our method can achieve the SOTA in closely interactive cases.

\begin{table}
    \begin{center}
        \resizebox{1.0\linewidth}{!}{
            \begin{tabular}{l|c c c c}
            \noalign{\hrule height 1.5pt}
            \begin{tabular}[l]{l}\multirow{1}{*}{Method}\end{tabular}
                &MPJPE &PA-MPJPE &MPVPE &Interaction  \\
            \noalign{\hrule height 1pt}
            Human4D~\cite{goel2023humans}   &72.1        &52.4            &88.6            &--      \\
            CLLIF~\cite{li2022cliff}   &91.3        &53.6            &109.6            &141.5      \\
            BEV~\cite{sun2022putting}   &91.8        &52.2            &101.2            &131.0      \\
            GroupRec~\cite{huang2023reconstructing}   &82.4        &51.6       &88.6         &98.8      \\
            BUDDI~\cite{muller2023generative}   &96.8        &70.6            &116.0            &102.6      \\
            \hline
            \textbf{Ours}  &\textbf{63.1} &\textbf{47.5} &\textbf{76.4} &\textbf{81.4} \\
            \noalign{\hrule height 1.5pt}
            \end{tabular}
        }
\vspace{-2mm}
\caption{\textbf{Comparisons on Hi4D.} Our method can outperform existing single-person and multi-person approaches in terms of joint position accuracy and interaction state. ``--" means the results are not available.}
\label{tab:Hi4D}
\end{center}
\vspace{-8mm}
\end{table}

We also conduct experiments on 3DPW datasets. We select all interactive sequences as a benchmark to evaluate our method. \cref{tab:3DPW} shows the results of baseline methods and ours. Although the data used for training the diffusion model are captured in indoor scenes, our method can also predict satisfactory results in outdoor scenarios. The results reveal that our method can achieve better human interactions than previous works.

We further conduct a qualitative comparison on CHI3D in \cref{fig:comparison}. Since BUDDI is a two-stage method, its performance strongly depend on the quality of 2D poses. However, the 2D pose detection always fails in close interactive scenarios. 
Consequently, it may produce incorrect body poses. In addition, BEV showes inter-person penetrations due to the lack of physical constraints. In contrast, our method can get better performance with the assistance of interaction prior and physical guidance.

\begin{table}
    \begin{center}
        \resizebox{1.0\linewidth}{!}{
            \begin{tabular}{l|c c c c}
            \noalign{\hrule height 1.5pt}
            \begin{tabular}[l]{l}\multirow{1}{*}{Method}\end{tabular}
                &MPJPE &PA-MPJPE &MPVPE &Interaction  \\
            \noalign{\hrule height 1pt}
            Human4D~\cite{goel2023humans}   &72.9        &49.1            &\textbf{107.0}            &--      \\
            BEV~\cite{sun2022putting}   &78.3        &\textbf{48.5}            &116.9            &136.4      \\
            GroupRec~\cite{huang2023reconstructing}   &73.3        &48.7            &109.4            &110.6      \\
            BUDDI~\cite{muller2023generative}   &83.6        &53.6       &126.1            &113.1      \\
            \hline
            \textbf{Ours}      &\textbf{70.6} &51.4 &107.9 &\textbf{100.3} \\
            \noalign{\hrule height 1.5pt}
            \end{tabular}
        }
\vspace{-2mm}
\caption{\textbf{Comparisons on 3DPW.} Our method can achieve competitive performance in terms of joint accuracy in outdoor scenarios. In addition, we still produce better interactions without training on in-the-wild video data.}
\label{tab:3DPW}
\end{center}
\vspace{-8mm}
\end{table}

\begin{figure*}
    \begin{center}
    \includegraphics[width=1.0\linewidth]{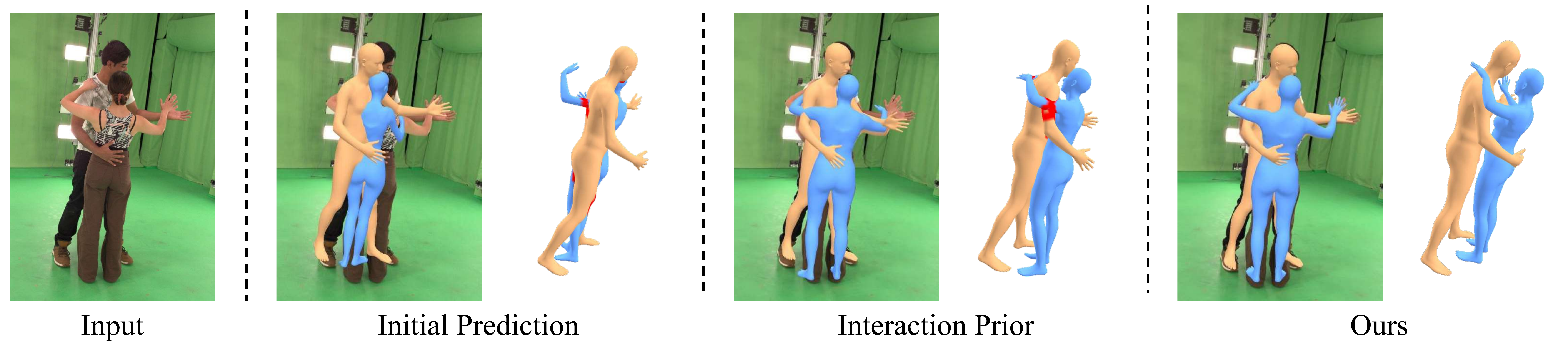}
    \end{center}
    \vspace{-7mm}
    \caption{Ablation study. The initial prediction is severely affected by visual ambiguity and cannot reconstruct natural interaction. The interaction prior update the motions from initial prediction with proxemic behaviours. Although the prior can produce better interaction, it still suffers from inter-person penetrations. With the distribution adaption, our method can refine the results, and reconstruct accurate interaction and physical contacts.}
\label{fig:ablation}
\vspace{-6mm}
\end{figure*}

\begin{table}
    \begin{center}
        \resizebox{1.0\linewidth}{!}{
            \begin{tabular}{l|c c c c}
            \noalign{\hrule height 1.5pt}
            \begin{tabular}[l]{l}\multirow{1}{*}{Method}\end{tabular}
                &MPJPE &PA-MPJPE &MPVPE &Interaction  \\
            \noalign{\hrule height 1pt}
            Initial Prediction   &73.7        &55.8            &88.4            &93.7      \\
            Single Branch   &64.1       &48.6        &81.5     &89.3       \\
            w/o prior   &67.6         &50.3        &83.2      &90.1       \\
            w/o physical guidance   &63.6         &\textbf{47.0}      &\textbf{76.1}             &86.1       \\
            \textbf{Ours}  &\textbf{63.1} &47.5 &76.4 &\textbf{81.4} \\
            \noalign{\hrule height 1.5pt}
            \end{tabular}
        }
\vspace{-3mm}
\caption{\textbf{Ablations on Hi4D.} "Single Branch" uses a single-person prior, and the two characters cannot share information during the inference. "w/o prior" and "w/o physical guidance" denote our model without the discrete prior and physical guidance.}
\label{tab:ablation}
\end{center}
\vspace{-9mm}
\end{table}

\subsection{Ablation Study}\label{sec:Ablation}
\noindent\textbf{Discrete interaction prior.} 
Reconstructing closely interactive humans from single-view images is a highly ill-posed problem due to the inter-person occlusions. The image can provide limited valid information for the reconstruction. The interaction prior models the interactive behaviours in discrete codebooks and can incorporate proxemic prior knowledge into the framework. For a noisy initial prediction, which may not follow the interactive behaviours, we find the closet interaction in the codebooks to refine the results. In \cref{fig:ablation}, we found that the incorrect predicted results can be adjusted with the prior, and the output of the prior can be a valid interaction to be used in the reverse diffusion process. Even for the severely occluded cases, the prior can still produce plausible 3D poses with the interaction relationships. In \cref{tab:ablation}, we found that the model performance decrease Without the prior.

\noindent\textbf{Physical guidance.} 
Although the prior can provide interactive behaviours to assist the reconstruction, the output results still suffer from penetrations. 
We thus provide a BVH method to detect colliding triangles and calculate the penetration loss with local distance field. Different from previous works that directly use the constraint as a supervision, we also incorporate the loss values as a condition in the reverse diffusion process. In each timestep, the model can sense the penetration and refine the results in the next step. In \cref{fig:ablation}, we can observe that the penetration is alleviated with the physical guidance. Although the model inference does not obey physical laws, the model can get feedback of the penetration with this strategy and thus produces better results compared to direct supervision.

\noindent\textbf{Dual-branch diffusion model.} 
To investigate the importance of interaction, we use a single-branch model to replace the dual-branch structure in discrete prior and diffusion model. In the prior, we use a single codebook to model human motions, which is similar to \cite{zhang2023t2m}. During the distribution adaption process, the initial motions is refined using the same codebook and not share information between two characters. Although it can also predict accurate relative joint and vertex positions, the interaction error is large due to the lack of proxemics.     
\vspace{-1mm}
\section{Limitation and Future Work}\label{sec:Limitation}
Although our method can reconstruct interactive humans from single-view videos, there still exist some limitations. First, the current design can only support two-person interactions. If the 
setting is extended to reconstruct social interactions with more people, 
our method has to maintain more codebooks and may result in redundancy. In the future, the network can be improved to aggregate similar motions in the same codebook for more general social interactions. Second, although we use in-the-wild image datasets to train the initial prediction network, our model performance may still 
deteriorate in outdoor scenarios. 
As a result, to build a in-the-wild dataset for close interaction is also a promising direction in the future. 

\vspace{-1mm}
\section{Conclusion}\label{sec:Conclusion}
In this work, we introduce a novel dual-branch diffusion model to incorporate proxemic behavior, physical constraint, and image observation to reconstruct close interactive humans. To alleviate the impact of occlusions and visual ambiguities, the proposed discrete prior encodes human close interactions in two codebooks and can provide additional knowledge for the reconstruction. We further formulate the reconstruction as a distribution adaption to consider the uncertainty of the ill-posed problem. Finally, the model refine the distribution under the guidance of observations in each reverse diffusion timestep, and can produce accurate and realistic interactions.

\vspace{-2mm}
\paragraph{Acknowledgement.} This research is supported by the
National Research Foundation, Singapore under its AI Singapore Programme (AISG Award No: AISG2-RP-2021-024) and China Scholarship Council under Grant Number 202306090192.

{
    \small
    \bibliographystyle{ieeenat_fullname}
    \bibliography{main}
}

\clearpage
\setcounter{page}{1}
\maketitlesupplementary

In the supplementary material, we first provide the details of VQ-VAE, motion representation, and training procedures to help the reproduction of the experimental results. More comparisons, analyses, and qualitative experiments are also conducted to further demonstrate the superiority of the proposed method. Finally, we show some failure cases to illustrate the limitations of the current method. We also provide a supplementary video for this work.

\section{VQ-VAE}\label{sec:VQVAE}
A naive training of VQ-VAE suffers from codebook collapse~\cite{razavi2019generating}. To avoid the limitation, we adopt exponential moving average (EMA) and codebook reset (Code Reset) to improve the codebook utilization. Specifically, EMA updates the codebook smoothly: $C^t \leftarrow \lambda C^{t-1}+(1-\lambda) C^t$, and $C^t$ is the codebook in the current iteration. $\lambda=0.99$ is a exponential moving constant. Code Reset finds inactivate codes during the training and reassigns them according to input data.

For the model architecture, the encoder of the discrete interaction prior consists of a motion embedding layer, a positional encoding layer, and 4 transformer blocks. The decoder has a motion decoding layer and 4 transformer blocks. Each codebook has a size of 256 $\times$ 256.

\section{Motion representation}\label{sec:Representation1}
Previous works~\cite{zhang2023t2m} always represent human motion in a canonical space, and the global rotation and translation are obtained by accumulating local angular and linear velocities. This representation cannot be directly applied to multi-person scenarios since it does not maintain the person-to-person spatial relationships. To address this problem, we use the root position of character $a$ in the first frame as origin and transform the interactive motions to the new coordinate. 
Consequently, the joint positions and velocities are kept in the world frame. In addition, the two-person interactions satisfy commutative property, which means the interaction $\left\{\mathbf{x}^{a}, \mathbf{x}^{b} \right\}$ and $\left\{\mathbf{x}^{b}, \mathbf{x}^{a} \right\}$ are equivalent.

\section{Implementation details}\label{sec:details}
The diffusion model has the same stracuture as the VQ-VAE encoder, which contains a motion embedding layer, a positional encoding layer, and 4 transformer blocks. The number of diffusion timesteps is set to 100 during the training stage. In the inference, we adopt DDIM sampling strategy~\cite{song2020denoising} with 5 timesteps to achieve the distribution adaption. For the projection loss gradients, we use ViT pose~\cite{xu2022vitpose} to predict 2D poses. To train the diffusion model, 25\% of projection loss gradients and image features are randomly masked. The frame length of interactive motions is 16, and the batch size for VQ-VAE and diffusion model are 256 and 32, respectively. All the models are trained with AdamW~\cite{loshchilov2017decoupled} optimizer using a learning rate of 1e-4 on a single GPU of NVIDIA GeForce RTX 4090.

\begin{table}
    \begin{center}
        \resizebox{0.6\linewidth}{!}{
            \begin{tabular}{c|c c}
            \noalign{\hrule height 1.5pt}
            \begin{tabular}[l]{l}\multirow{1}{*}{Dataset}\end{tabular}
                &w/o proj. loss &Ours  \\
            \noalign{\hrule height 1pt}
            3DPW   &73.8         &70.6              \\
            Hi4D  &64.2    &63.1  \\
            \noalign{\hrule height 1.5pt}
            \end{tabular}
        }
\vspace{-3mm}
\caption{\textbf{Ablation on projection loss gradients.} ``w/o proj. loss" denotes our model without the projection loss gradients. The number are MPJPE.}
\label{tab:loss}
\end{center}
\vspace{-6mm}
\end{table}

\begin{table}
    \begin{center}
        \resizebox{0.6\linewidth}{!}{
            \begin{tabular}{c|c c}
            \noalign{\hrule height 1.5pt}
            \begin{tabular}[l]{l}\multirow{1}{*}{Method}\end{tabular}
                &Accel$\downarrow$ &A-PD$\downarrow$  \\
            \noalign{\hrule height 1pt}
            GroupRec   &25.2         &1.34              \\
            Ours  &10.7    &1.15  \\
            \noalign{\hrule height 1.5pt}
            \end{tabular}
        }
\vspace{-3mm}
\caption{Accel and A-PD are acceleration error and average penetration depth, respectively.}
\label{tab:Accel}
\end{center}
\vspace{-6mm}
\end{table}

\begin{figure*}
    \begin{center}
    \includegraphics[width=1.0\linewidth]{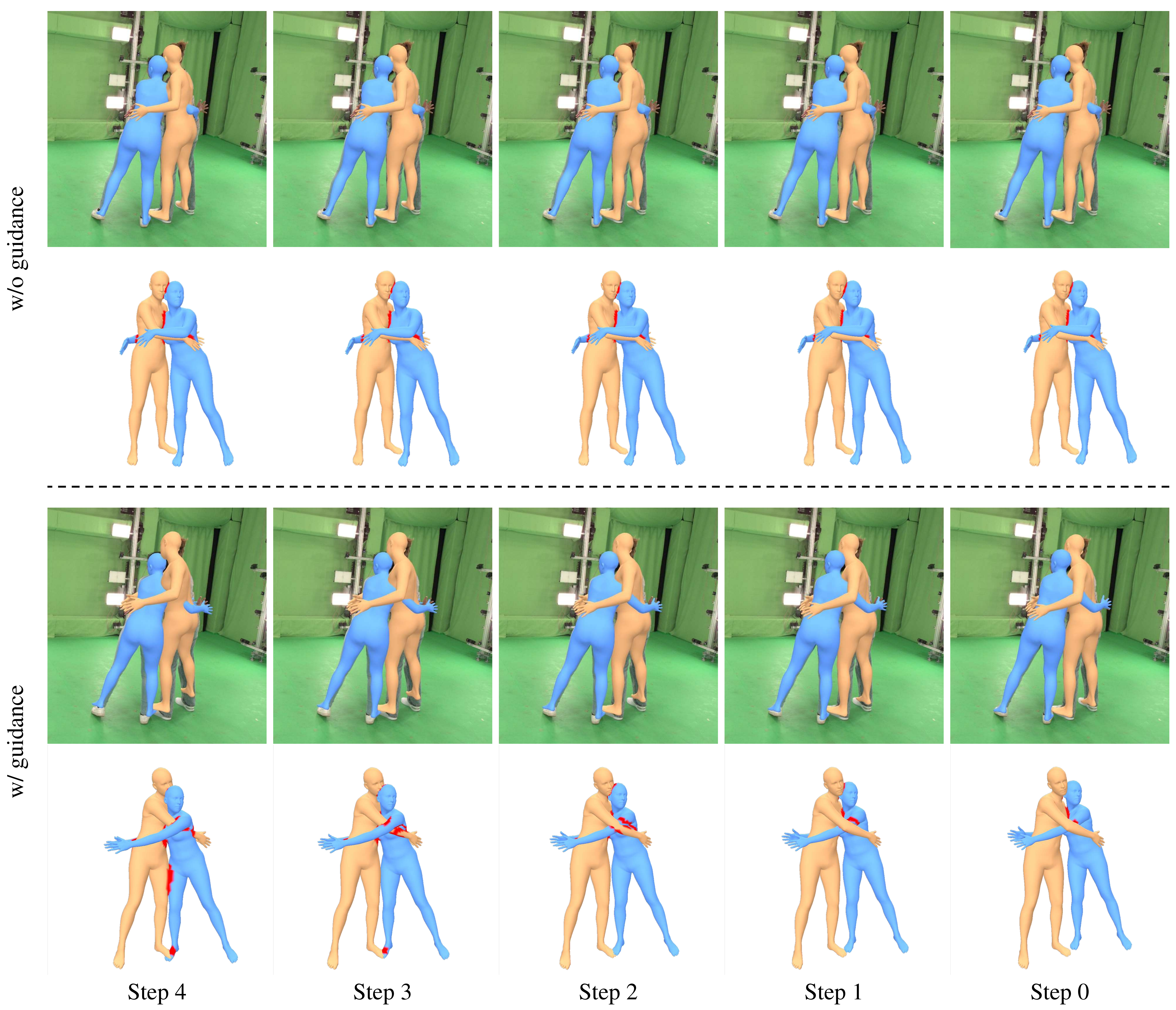}
    \end{center}
    \vspace{-6mm}
    \caption{Comparison between the models with and without the guidance. Without the guidance, the reverse diffusion process has limited effect in improving the interations.}
\label{fig:intermediate}
\vspace{-4mm}
\end{figure*}

\begin{table}
    \begin{center}
        \resizebox{1.0\linewidth}{!}{
            \begin{tabular}{c|c c c c}
            \noalign{\hrule height 1.5pt}
            \begin{tabular}[l]{l}\multirow{1}{*}{Timesteps}\end{tabular}
                &step=1 &step=3  &step=5 &step=10 \\
            \noalign{\hrule height 1pt}
            MPJPE   &68.2         &64.4    &63.1   &63.0           \\
            \noalign{\hrule height 1.5pt}
            \end{tabular}
        }
\vspace{-3mm}
\caption{\textbf{Ablation on number of duffusion timesteps.} The ablation is conducted on Hi4D.}
\label{tab:timestep}
\end{center}
\vspace{-6mm}
\end{table}

\begin{figure*}
    \begin{center}
    \includegraphics[width=1.0\linewidth]{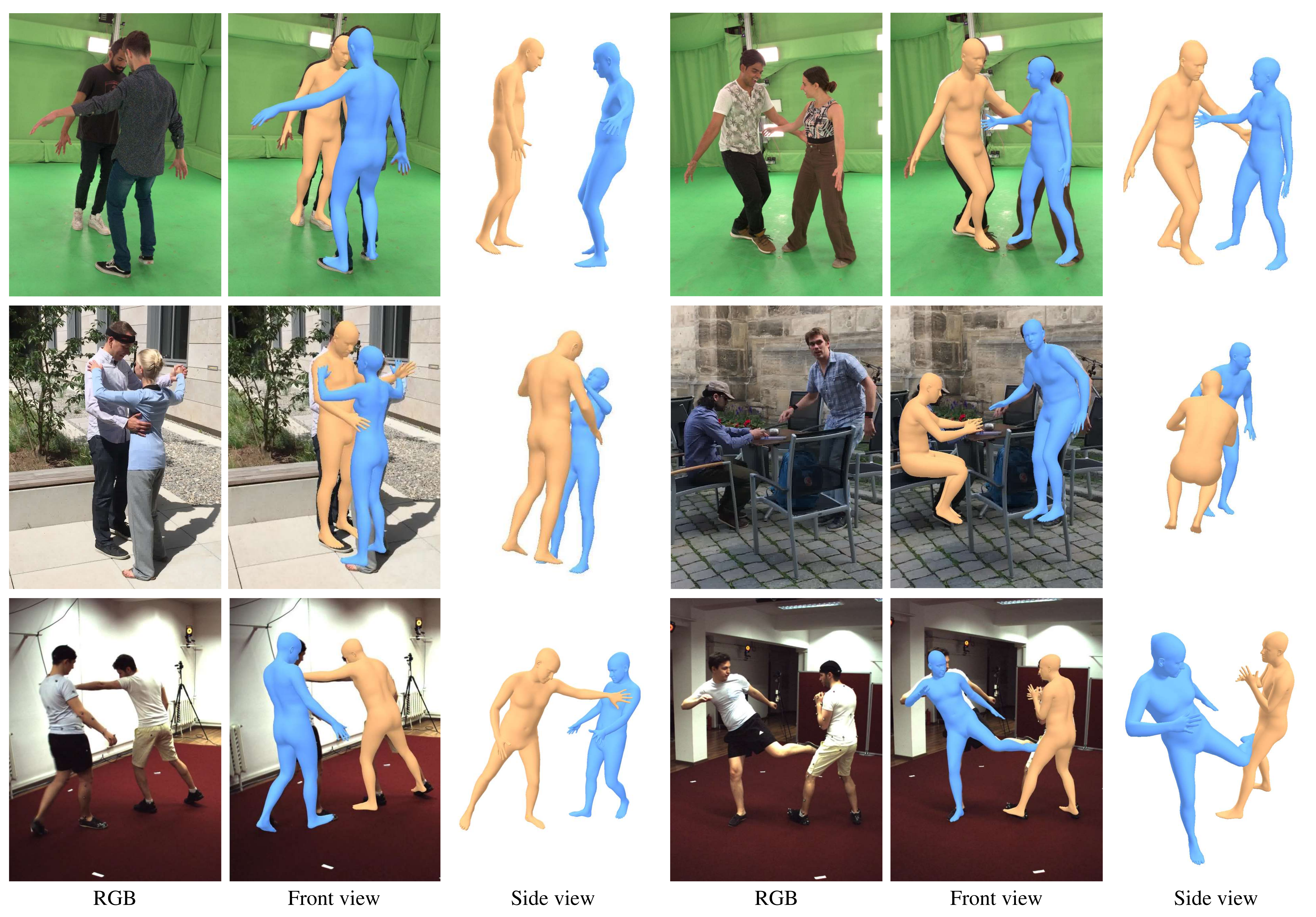}
    \end{center}
    \vspace{-6mm}
    \caption{More qualitative results on Hi4D, 3DPW and CHI3D datasets. Our method can reconstruct closely interactive humans with plausible body poses, natural proxemic relationships and accurate physical contacts from single-view inputs}
\label{fig:more_results}
\vspace{-4mm}
\end{figure*}

\section{Extended experiments}\label{sec:experiments}
\noindent\textbf{Projection loss gradients.} We also investigate the projection loss gradients, which provide signals to guide the reverse diffusion process and enforce the 3D models to be consistent with image observations. The term can improve joint accuracy for common scenarios. However, it has limited effect on severely occluded cases since the current state-of-the-art 2D pose detectors cannot produce reliable results for ambiguous images. In \cref{tab:loss}, Hi4d has more complex interactions than 3DPW, and the performance on 3DPW dataset significantly decreases without the projection loss gradients. 

\noindent\textbf{Diffusion guidance.} In \cref{fig:intermediate}, we show the intermediate results in each timestep during the inference phase. The inference contains 5 timesteps with the denoising diffusion implicit models~\cite{song2020denoising}. Although the reverse diffusion has the same timesteps, we find that the denoising model without the guidance produces slight changes and the final results still show severe penetrations. In contrast, our model can refine the interactions and alleviate penetrations, which demonstrates the importance of proposed guidance.

\noindent\textbf{Diffusion timesteps.} We analyze the impact of different timesteps in \cref{tab:timestep}. The performance increases with more timesteps at first and then becomes stable. To balance the accuracy and efficiency, we use 5 timesteps in the inference phase.

\noindent\textbf{Penetration and acceleration error.}
In \cref{tab:Accel}, we further use the average penetration depth (A-PD)~\cite{rong2021monocular}, which reflects the degrees of inter-penetration, to evaluate the body contact and penetration, and our method can produce better performance. Our method also outperforms GroupRec~\cite{huang2023reconstructing} on the acceleration error due to the proposed velocity loss and temporal architecture.

\noindent\textbf{More results.} We also show more qualitative results in \cref{fig:more_results}. Our method can reconstruct closely interactive humans with plausible body poses, natural proxemic relationships and accurate physical contacts from single-view inputs.

\end{document}